# Segmentation Criteria in the Problem of Porosity Determination based on CT Scans

V. Kokhan[1, 2], M. Grigoriev[3], A. Buzmakov[4], V. Uvarov[5], A. Ingacheva[1, 2], E. Shvets[1], M. Chukalina[2, 4]

[1]Institute for Information Transmission Problems (Kharhevich Institute) Russian Academy of Sciences, 127051, Moscow, Russia
[2]Smart Engines, 117312, Moscow, Russia
[3]Institute of Microelectronics Technology and High-Purity Materials Russian Academy of Sciences, 142432, Chernogolovka, Russia
[4]FSRC "Crystallography and Photonics" Russian Academy of Sciences, 119333, Moscow, Russia
[5]Merzhanov Institute of Structural Macrokinetics and Materials Science
Russian Academy of Sciences, 142432, Chernogolovka, Russia

## ABSTRACT

Porous materials are widely used in different applications, in particular they are used to create various filters. Their quality depends on parameters that characterize the internal structure such as porosity, permeability and so on. Computed tomography (CT) allows one to see the internal structure of a porous object without destroying it. The result of tomography is a gray image. To evaluate the desired parameters, the image should be segmented. Traditional intensity threshold approaches did not reliably produce correct results due to limitations with CT images quality. Errors in the evaluation of characteristics of porous materials based on segmented images can lead to the incorrect estimation of their quality and consequently to the impossibility of exploitation, financial losses and even to accidents. It is difficult to perform correctly segmentation due to the strong difference in voxel intensities of the reconstructed object and the presence of noise. Image filtering as a preprocessing procedure is used to improve the quality of segmentation. Nevertheless, there is a problem of choosing an optimal filter. In this work, a method for selecting an optimal filter based on attributive indicator of porous objects (should be free from "levitating stones" inside of pores) is proposed. In this paper, we use real data where beam hardening artifacts are removed, which allows us to focus on the noise reduction process.

## 1. INTRODUCTION

Porous structured materials are widely used in industry. Their quality depends on such parameters as porosity, permeability, and so on. However, direct measurement of these parameters requires an invasive intervention in a material that can lead to its destruction. For this reason non-invasive methods for obtaining images are used to solve a parameters estimation problem. The most common of these non-invasive methods is computed tomography (CT). After segmentation of the reconstructed gray images the characteristics can be evaluated [1]. We will use a binary segmentation, which is usually concerned with an image model evaluation by the image itself [2]. Traditional intensity threshold approaches are not reliably produce quantitatively accurate. A machine learning based pixel classification produced significantly improved quality segmentations [1, 3] but requires big training data sets. It is known that the image of an object obtained with an optical device is usually distorted by noise [4]. The presence of noise in images is the major obstacle to segment these images of porous structures. In CT, noise can have various causes. Noise appears due to an insufficient number of radiation quanta or thermal noise of a detector [5, 6]. In addition to noise, CT images may also contain artifacts caused, for example, by an insufficient amount of projection data or properties of reconstruction algorithms [7]. Thermal shift of the X-ray source during the experiment can produce the problems. In this paper, we use real data where some of the artifacts like beam hardening [8] are removed, and the remaining ones are low-grade, which allows us to focus on the noise reduction process.

In image processing, the smoothing filters, such as Gaussian and median, are traditionally used to reduce noise. Such filters are effective in many cases however they do not take into account the structure of a filtered object and therefore blur not only noisy areas, but also semantic boundaries of the object [9] and can also lead to artifacts [10], what can lead to incorrect assessment of the pore parameters. Thus, such filters do not satisfy us. In recent years, it has often been noted in the literature that filtering with a preservation of boundaries is necessary for images of porous structures. The following noise reduction algorithms are often used for denoising images of porous structures: median filtering in combination with unsharp mask sharpening [11-13], an anisotropic diffusion filter [9, 14-16], bilateral filtering [17-21]. It should be noted that guided filtering [22, 23] also belongs to such algorithms, but at present it is extremely rarely used in the problem under consideration. It is also worth noting that the combination of the median filter and unsharp masking can lead to significant image distortion, since after applying the median filtering some semantic boundaries can be lost without the possibility of restoring them with unsharp mask sharpening.

Thus, we can conclude that the problem of filtering images of porous structures is relevant today, but the criteria to choose an optimal method for filtering is unknown. The purpose of this work is to develop a method for selecting a filter type and its parameters in order to improve the quality of images of porous structures. Since the considered algorithms do not take into account the nature of the appearance of noise on the tomographic images of the porous structure, and there are no rules for choosing the optimal parameters of these filters, in this paper we examine the application of 4 types of filters to the porous structure and propose a choice of the filter type and its parameters based on the results of segmentation. After the segmentation process, structural elements appear that they do not have any connection with the bulk of the material. We call such elements "levitating stones". By changing the filter parameters we will change the number of "levitating stones" but in the same time we blur or sharp the borders. The optimal set of parameters is the set with minimal number of "levitating stones" and preserved boundaries on the segmented image. At the end of the article, we propose a method that decides the fate of the rest of "levitating stones" (fix to the balk or ignor).

## 2. FILTERS PARAMETERS SELECTION METHOD

### 2.1 Filters

We consider the following types of filters: median, anisatropic diffusion, bilateral and guided filters. In this section, we will explain the basic principles of their work.

The **median** filter is widely known in the field of image processing and is widely used today. The median filter for image $I$ and window $W$ can be written as a sequence of three steps, that applied to all image points: choose values that fell into the filter window $W$. The central position of the window point match with the current image point; sort window values; select the median value of sorted values and record it to the image point. The parameters of median filter are window height $h$ and window width $w$.

The **anisotropic** diffusion filter is the iterative image filtering [24]. Image blurring is presented as a diffusion process defined by the following expression:

$$I_{N+1} = I_N + \lambda \sum_{NSEW} g(\|\nabla I\|)\nabla I, \quad (1)$$

where $I_N$ denotes the image $I$ at $N$th iteration; $\lambda$ is diffusion speed coefficient; $NSEW$ is the four directions; $\nabla I$ is image gradient; $g(|\nabla I|)$ is diffusion coefficient.

The diffusion coefficient proposed in the original work has the following form:

$$g(\nabla I) = \frac{1}{\left(1+\left(\frac{|\nabla I|}{K}\right)^2\right)}, \quad (2)$$

where $K$ is a user-defined parameter. Such a filter was called an anisotropic diffusion filter. It preserves boundaries in areas where the gradient is large. Such areas usually correspond to the boundaries in the image. The disadvantage of this filter is iterative evaluation, which significantly affects the speed of the algorithm [23]. The parameters of anisotropic diffusion filter are $K$, $\lambda$ and number of iteration ($N$).

The **bilateral** filter uses both the spatial position information of pixels and the pixel luminance value. The equation for calculating a filer window $W^{bf}$ can be written as follows:

$$W_{ij}^{bf}(I) = k_i^{-1} \exp\left(-\frac{|x_i-x_j|^2}{\sigma_s^2}\right) \exp\left(-\frac{|I_i-I_j|^2}{\sigma_r^2}\right), \qquad (3)$$

where $k_i$ is a normalizing parameter to ensure that $\sum_j W_{ij}^{bf} = 1$. The parameters $\sigma_s$ and $\sigma_r$ regulate the spatial similarity and the range (intensity/color) of the similarity respectively. The window size (height $h$ and width $w$) is also filter parameters.

The **guided** filter is also window filter [24]. The equation for calculating a window $W^{gf}$ can be written as follows:

$$W_{ij}^{gf}(I) = \frac{1}{|\omega|^2} \sum_{k:(i,j) \in \omega_k} \left(1 + \frac{(I_i-\mu_k)(I_j-\mu_k)}{\sigma_k^2+\varepsilon}\right), \qquad (4)$$

where $\omega_k$ is a window centered at the pixel $k$; $\mu_k$ and $\sigma_k^2$ are the mean and variance of $I$ in $\omega_k$. Using the guided filter, evaluation time does not depend on the size of the window, in contrast to the bilateral filter, which makes this filter potentially the fastest of the considered filters that preserve boundaries.

## 2.2 Experiment

We used the filter implementations available in Python open access libraries: median filter from SciPy, anisatropic diffusion filter from MedPy, bilateral filter and guided filter from OpenCV. The filters were studied on experimental data obtained as a result of tomographic reconstruction of an object, which is a porous cermet membrane based on coarse-grained aluminum oxide with an average particle size of 250-350 microns. The research sample was made at the Institute of Structural Macrokinetics and Problems of Materials Science A.G. Merzhanova RAS (Chernogolovka). The porous structure has been measured with laboratory tomographic set-up with a 9-micron pixel size of FRC "Crystallography and Photonics" RAS [26]. Grey images have been reconstructed with own software [26]. The representative volume of the object was evaluated in [27], it is $250 \times 250 \times 250$ pixels.

After the segmentation by unbalanced Otsu method [28], structural elements appear that they do not have any connection with the bulk of the material ("levitating stones"). Usually they have relatively small sizes about one voxel (Fig.1). Their occurrence can be caused by the fact that the substance connecting them with the bulk of the material could have a relatively low absorption coefficient, and could be classified as pores, but "levitating stones" can also appear due to significant noise. The latter require special attention. Regardless of the nature of the occurrence, such stones will distort the numerical characteristics of the porous structure.

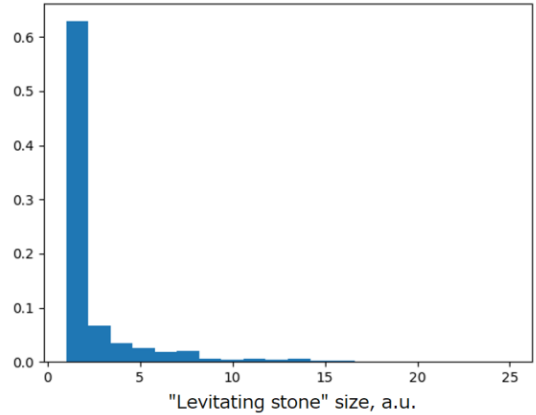

Figure 1. Histogram of size distribution of levitated stones

By unbalanced Otsu method one divide the histogram of image intensity into two classes proposing that these two classes are realizations of normal distributions with equal variances but different weights. In Fig. 2, one slice and its corresponding segmented image are shown.

Since the number of one-voxel stones is greater than the number of larger stones, we assume that one-voxel stones are due to noise and therefore the number of one-voxel stones is a good estimate of the noise level. Thus, we intend to use minimization of the amount of one-voxel "levitating stones" as a method of choose filter parameters. However, such a method requires a stopping criteria. We think that acceptable image distortion should be such a criteria. Distortion will be measured by the normalized Euclidean metric.

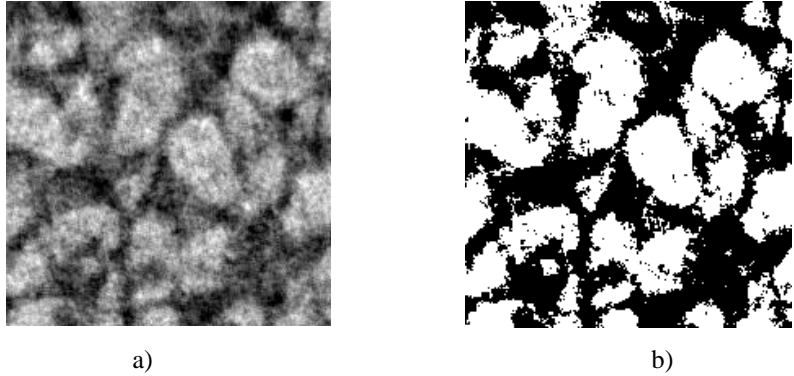

a) b)

Figure 2. a) One of the reconstructed slices, b) corresponding segmented image with "levitating stones"

$$\Delta = \frac{\|I_{original} - I_{filtered}\|_2}{V} = \frac{\sqrt{(I_{original} - I_{filtered})^2}}{V}, \quad (5)$$

where $I_{original}$ and $I_{filtered}$ are images before and after filtering; $V$ is the number of voxels in the image.

As the reference filter we chose the median filter as the most common in solving the observed problem, and also being classic in the field of image processing. The images intensity range from 0 to 255. For a median filter with a linear window size w = 3, distortion is $\Delta$ = 0.0072. With this value, the influence of filtering on the object structure distortion is already noticeable, and for w = 5 ($\Delta$ = 0.0096) this effect is already obvious (Fig. 3). For this reason, we chose 0.0072 as the acceptable image distortion.

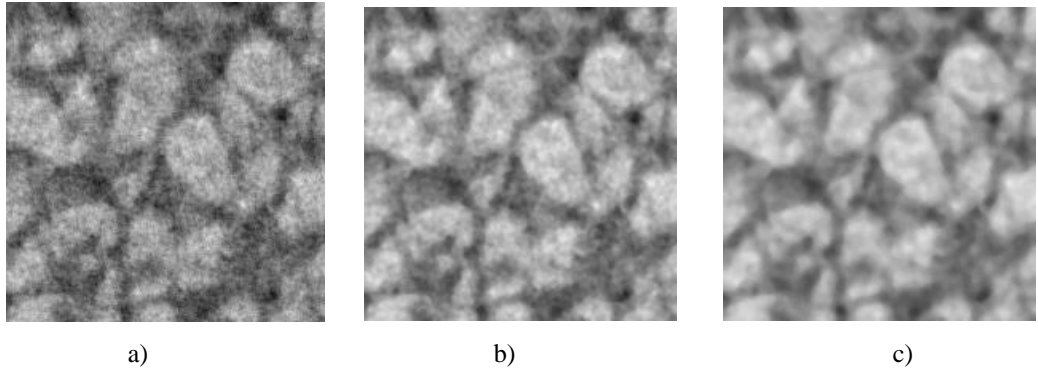

a) b) c)

Figure 3. Median filter usage: a) original image, b) filtered with linear window size $w = 3$, c) linear window size $w = 5$

The results of the selection of optimal filter parameters are shown in Table 1, slices of the obtained segmented images are shown in Fig. 4. From Table 1, it can be seen that the number of one-voxel stones is minimal when using an anisotropic diffusion filter.

Table 1. Experimental results

| Filter | Optimal parameters | Number of one-voxel stones | Distortion value |
|---|---|---|---|
| Without filtration | - | 473 | 0.0 |
| Median | $h = 1, w = 3$ | 9 | 0.0072 |

| Anisatropic diffusion | $N = 8; \lambda = 0.2; K = 20$ | 3 | 0.0072 |
|---|---|---|---|
| Bilateral | $h = 1, w = 7; \sigma_{color} = 0,5; \sigma_{space} = 1.3$ | 7 | 0.0072 |
| Guided | $w = 3; \sigma_k = 0.275$ | 5 | 0.0072 |

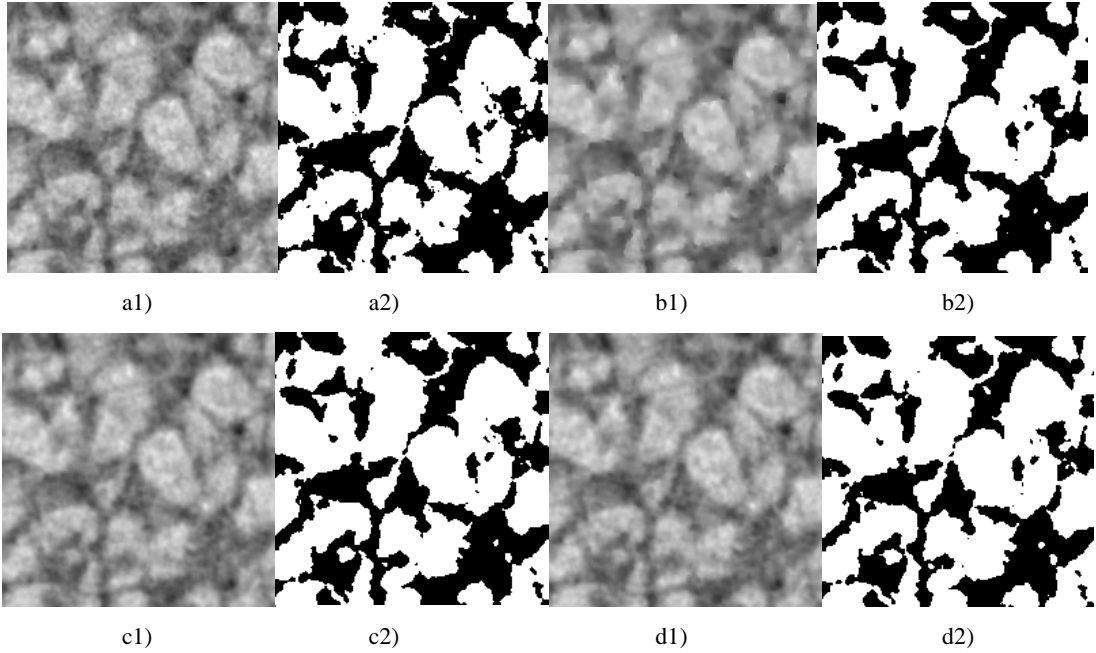

a1)     a2)     b1)     b2)

c1)     c2)     d1)     d2)

Figure 4. After filtering (left – grey image, right – segmented image)

a) median filter, b) anisotropic diffusion filter, c) bilateral filter, d) guided filter

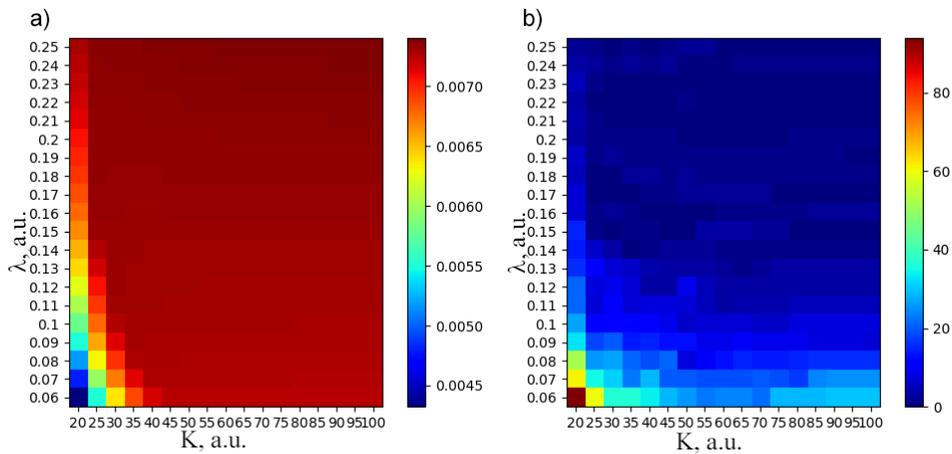

Figure 5. Dynamics of image blur depending on the filter parameters (a) and dependence of the number of "levitating stones" on the filter parameters (b) for anisotropic diffusion filter

To illustrate the dynamics of image blur depending on the filter parameters used, we calculated the standard deviation of the filtering result from the unfiltered image for anisotropic diffusion filter (Fig. 5a). Figure 5b presents the dependence of the number of "levitating stones" on the filter parameters for anisotropic diffusion filter. We can see that in the region $\lambda \sim 0.2, K \sim 20$ we have number of "levitating stones" close to minimum while the edges of the pores are not yet destroyed.

## 3. POST-PROCESSING

To process the rest of "levitating stones" we propose to use a relative distance between the stone and the object. If the stone appears near the object and its volume is relatively large in comparison with the pore diameter, then there is a high probability that it is a part of the object separated after incorrect segmentation, which can be attached to the object, but if the small stone is far from the object, then it probably refers to background or noise and therefore, to an undesirable artifact. The metric we offer has the following form:

$$\hat{d} = \frac{d}{\sqrt[3]{V_s}}, \tag{6}$$

where $d$ is the shortest Euclidean distance from the stone to the object; $V_s$ is the number of voxels in the stone. We propose that if the metric value is above a given threshold, then the stone is actually levitated and needs to be removed. If the metric value is below the threshold, then the stone should be attached to the bulk.

## 4. CONCLUSION

In this paper, the method for selecting a filter and its parameters is proposed for the objects with porous structures. The method is based on the specific parameters of porous objects, namely the number of one-voxel "levitating stones". It is taken into account that the filtration should not strongly distort the object, thus we used the boundary preserving filter, namely anisatropic diffusion, bilateral and guided filters. The study was carried out on the real CT data. The study showed that for the given object an anisatropic diffusion filter shows better results. A post-processing method was also presented To process the rest of "levitating stones".

This work was partly supported by Russian Foundation for Basic Research (project 18-29-26019 and 18-29-26037).